\documentclass[sigconf,nonacm]{acmart}

\AtBeginDocument{%
  }

\setcopyright{none}
\copyrightyear{2025}
\acmYear{2025}

\acmConference[KDD SKnowLLM Workshop '25]{KDD Workshop on Structured Knowledge for Large Language Models}{2025-08-04}{Toronto, Canada}
\acmISBN{978-1-4503-XXXX-X/18/06}

\usepackage{hyperref}       
\usepackage{url}            
\usepackage{booktabs}       
\usepackage{amsfonts}       
\usepackage{nicefrac}       
\usepackage{microtype}      
\usepackage{xcolor}         

\usepackage{enumitem}
\setlist[itemize]{leftmargin=2em}
\usepackage{amsmath}
\usepackage{multirow}
\usepackage{array}
\usepackage{makecell}
\usepackage{rotating}  
\usepackage{graphicx}
\usepackage{wrapfig}  
\usepackage[breakable]{tcolorbox}
\usepackage{float}  
\usepackage{mdframed}
\usepackage{listings}

\begin{document}

\title{RASL: Retrieval Augmented Schema Linking for Massive Database Text-to-SQL}

\makeatletter
\renewcommand\@formatdoi[1]{\ignorespaces}
\makeatother

\author{Jeffrey Eben}
\email{jeffeben@amazon.com}
\affiliation{%
  \institution{Amazon}
  \city{Jersey City}
  \state{NJ}
  \country{USA}
}

\author{Aitzaz Ahmad}
\email{aitzaza@amazon.com}
\affiliation{%
  \institution{Amazon}
  \city{Seattle}
  \state{WA}
  \country{USA}
}

\author{Stephen Lau}
\email{lausteph@amazon.com}
\affiliation{%
  \institution{Amazon}
  \city{Seattle}
  \state{WA}
  \country{USA}
}

\thanks{Paper accepted to the KDD Workshop on Structured Knowledge for Large Language Models (SKnowLLM '25), Toronto, Canada}

\begin{abstract}
Despite advances in large language model (LLM)-based natural language interfaces for databases, scaling to enterprise-level data catalogs remains an under-explored challenge. Prior works addressing this challenge rely on domain-specific fine-tuning—complicating deployment—and fail to leverage important semantic context contained within database metadata. To address these limitations, we introduce a component-based retrieval architecture that decomposes database schemas and metadata into discrete semantic units, each separately indexed for targeted retrieval. Our approach prioritizes effective table identification while leveraging column-level information, ensuring the total number of retrieved tables remains within a manageable context budget. Experiments demonstrate that our method maintains high recall and accuracy, with our system outperforming baselines over massive databases with varying structure and available metadata. Our solution enables practical text-to-SQL systems deployable across diverse enterprise settings without specialized fine-tuning, addressing a critical scalability gap in natural language database interfaces.
\end{abstract}


\maketitle

\section{Introduction}
Text-to-SQL systems translate natural language questions into executable SQL queries, enabling non-technical users to extract insights from databases without SQL expertise. While these systems have evolved from rule-based approaches to powerful large language model (LLM) solutions \cite{dailsql}, scaling them to industrial settings with massive database catalogs remains an underexplored challenge.

Current state-of-the-art methods primarily leverage LLMs using techniques like task decomposition and prompt optimization, avoiding the overhead of maintaining fine-tuned models \cite{survey, mac_sql, mcs_sql, chase_sql}. However, these approaches face critical limitations when applied to enterprise environments with thousands of tables and tens of thousands of columns. In such massive database settings, providing comprehensive schema context to LLMs becomes untenable due to token limitations, computational costs, and semantic overload \cite{chess, dbcopilot, crush}. For example, a typical enterprise data catalog with 10,000 tables averaging 50 columns each would require over 500,000 schema entities—far exceeding current LLM context windows and creating prohibitive costs for commercial API usage.

Existing solutions for scaling to massive catalogs either rely on hierarchical selection methods that require domain-specific training and well-defined database-table hierarchies \cite{dbcopilot}, employ computationally intensive multi-agent frameworks that process entire schemas through many LLM calls \cite{chess}, or use complex optimization techniques that don't scale to truly massive schemas \cite{jar}. These methods struggle in real-world deployments where database architectures often follow monolithic NoSQL paradigms, metadata about join relationships is incomplete, or database schema is often changing. The scalability challenge becomes exponentially worse as database size increases: while methods may work reasonably well on benchmarks with hundreds of tables, they fail to maintain acceptable performance and cost efficiency when scaled to enterprise catalogs with thousands of tables \cite{chess}.

We present Retrieval Augmented Schema Linking (RASL), a novel approach designed specifically for text-to-SQL over massive database schemas without requiring fine-tuning or well defined database relations. RASL decomposes schemas into semantic entities, indexes them in a vector database, and employs a multi-stage retrieval process with relevance calibration to efficiently narrow the search space while maintaining compatibility with hosted LLM services.

Our contributions include: \begin{itemize} 
\item A zero-shot schema linking architecture that scales to massive databases with minimal preprocessing, no model training, and no requirement of known database hierarcy and join relations
\item An entity-level decomposition strategy with keyword-based context retrieval and entity-type relevance calibration
\item Empirical evidence of RASL's effectiveness on industrial-scale benchmarks
\end{itemize}

Our work bridges the gap between academic text-to-SQL research and industrial requirements, providing a practical solution for natural language interfaces to massive data environments.

\section{Related Works} Schema linking—mapping natural language elements to database components—becomes exponentially more challenging as database size increases. While some works have found schema-linking to be unnecessary on standard benchmarks with the latest foundation model offerings \cite{death_schema_linking}, others have shown that text-to-SQL performance degrades as database size increases and state-of-the-art methods are unable to scale to full industry-scale data catalogs \cite{chess}.

Several approaches have been proposed for massive database environments, each with limitations in industrial settings. DBCopilot \cite{dbcopilot} models schema linking as path generation through a hierarchical graph, first predicting the database, then tables, before generating SQL. While effective with clear database boundaries and known join relations, this approach struggles in monolithic data lake settings and requires extensive training on synthetic question-schema pairs, making adaptation to schema changes difficult. CHESS \cite{chess} incorporates an Information Retriever and Schema Selector to retrieve and prune context, but faces severe scalability issues with industry-sized datasets. It processes the full schema via many LLM calls, only using retrieval to augment already-identified schema elements with additional indexed context.  

CRUSH4SQL \cite{crush} embeds column names during build time and hallucinates a candidate schema from input questions to retrieve relevant columns. While conceptually similar to our approach, CRUSH is limited to column name schema context, showing degraded performance on complex benchmarks containing additional context such as descriptions and value formats. Additionally, its reliance on LLMs to hallucinate schemas for querying can lead to misalignment with ground truth database schemas where columns follow non-standard naming conventions.

RASL addresses these limitations through a zero-shot architecture that leverages both column-level and table-level context without requiring database hierarchy knowledge or comprehensive table relationship metadata. By separating build-time schema decomposition from inference-time retrieval, RASL provides an effective balance between accuracy and efficiency for industrial deployments where schemas frequently evolve across diverse storage paradigms.

\section{Preliminaries}
We begin by establishing notation for the components of our approach.

\paragraph{Database Schema} We define a database schema $S$ as a collection of tables $T = \{t_1, t_2, \ldots, t_n\}$. Each table $t_i$ consists of a set of columns $C_i = \{c_{i,1}, c_{i,2}, \ldots, c_{i,m_i}\}$, where $m_i$ is the number of columns in table $t_i$. The complete set of columns across all tables is denoted as $C = \cup_{i=1}^{n} C_i$.

\paragraph{Entity Types} We define $\Lambda = \{\lambda_1, \lambda_2, \ldots, \lambda_l\}$ as the set of all entity types, where each $\lambda_j$ represents a specific type of schema information (e.g., table description). Entity types are partitioned into table-level types $\Lambda_T \subset \Lambda$ and column-level types $\Lambda_C \subset \Lambda$.

\paragraph{Schema Entities} For each table $t_i$ and entity type $\lambda_j \in \Lambda_T$, we define entity $e(t_i, \lambda_j)$ as the representation of table $t_i$ according to type $\lambda_j$ (e.g., its name or description). Similarly, for each column $c_{i,k}$ and entity type $\lambda_j \in \Lambda_C$, we define entity $e(c_{i,k}, \lambda_j)$ as the representation of column $c_{i,k}$ according to type $\lambda_j$. The complete set of all entities is denoted as $E$.

\paragraph{Vector Representations} Each entity $e \in E$ is embedded in a $d$-dimensional space using an embedding function $\phi: E \rightarrow \mathbb{R}^d$. Similarly, a natural language question $q$ is embedded as $\phi(q) \in \mathbb{R}^d$. The similarity between two embeddings is measured using cosine similarity.

\section{Methodology}

\begin{figure*}[t]
    \centering
    \includegraphics[width=1.0\textwidth]{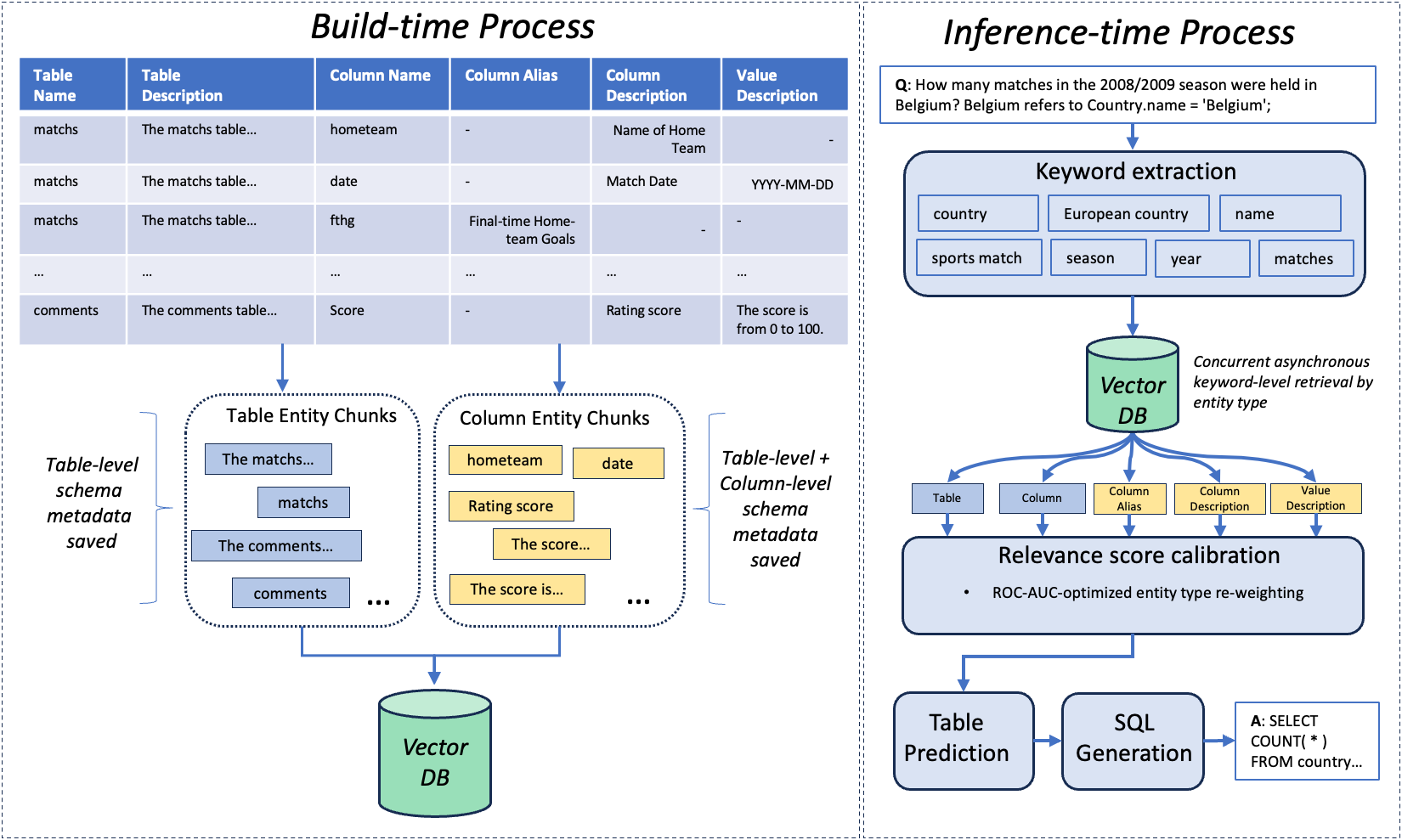}
    \caption{System overview. (left) Build-time process of constructing the schema metadata knowledge base. (right) Inference-time retrieval process for text-to-SQL applications.}
    \label{fig:system_overview}
\end{figure*}

\subsection{Overview} RASL addresses the challenge of scaling text-to-SQL to massive database catalogs through a two-phase approach: build-time knowledge base construction and inference-time retrieval augmented schema linking. Figure~\ref{fig:system_overview} illustrates our pipeline.

At build time, RASL decomposes database schema $S$ into semantic entities $E_{\Lambda_{T}}$ and $E_{\Lambda_{C}}$, which are embedded and indexed in a vector database with metadata tags incorporating full schema context for later reconstruction. At inference time, given a natural language question $q$, RASL extracts keywords $K$ and performs parallel retrieval for each $k \in K \cup \{q\}$ across each entity type $\lambda_j \in \Lambda$. For each retrieval query, RASL applies entity-type-level calibration to account for variably discriminative entity types when training samples are available. RASL then filters entities to retain only those belonging to the top $N$ tables, considering both table-level and column-level entities for table ranking. This filtered subset serves as input for LLM-based table prediction to identify the most relevant tables for the query. Finally, RASL loads the complete schema context for these predicted tables to support downstream SQL generation.

\subsection{Knowledge Base Construction}

\textbf{Schema Entity Decomposition}. We decompose database schema $S$ into semantic entities at table and column levels as defined in our preliminaries. Specific $E$ vary by dataset, with examples of $E_{\Lambda_T}$ including table names, aliases, and descriptions and examples of $E_{\Lambda_C}$ containing column names, aliases, descriptions, and value format descriptions. For example, consider a table \texttt{student\_club.member} with columns \texttt{first\_name}, \texttt{last\_name}, and \texttt{zip}. RASL would create separate entities: $e(\texttt{student\_club.member}, \lambda_{\text{table name}}) = $ "student\_club.member", $e(\texttt{first\_name}, \lambda_{\text{column name}}) = $ "first\_name", $e(\texttt{last\_name}, \lambda_{\text{column name}}) = $ "last\_name", and $e(\texttt{zip}, \lambda_{\text{column name}})$ $ = $ "zip". Each entity $e \in E$ is indexed with metadata tags linking it to the original schema structure, preserving hierarchical relationships for inference-time schema re-construction.

\textbf{Vector Embedding and Indexing}. Each semantic entity $e \in E$ is embedded using the embedding function $\phi$ to capture its semantic meaning in a $d$-dimensional vector space. These embeddings are indexed in a vector database optimized for similarity search, enabling efficient retrieval without task-specific fine-tuning.

\textbf{Table description Synthesis}. For tables with limited or missing descriptions, which is common across all datasets evaluated, we explore synthesizing descriptive text using an LLM that analyzes table structure, column names, and available metadata. This description is saved as $\lambda_{\text{table descr.}} \in \Lambda_T$, with specific details on synthesis prompts provided in \ref{app:table_synth}. For fair comparison with baselines, we primarily evaluate our system without the inclusion of $E_{\lambda_{\text{table descr.}}}$, with ablation studies exploring the effect of adding additional synthesized context on RASL's performance.

\subsection{Retrieval-Augmented Schema Linking}

\subsubsection{Question Decomposition} Inspired by CHESS-SQL \cite{chess}, we decompose each user question $q$ into keywords $K = \{k_1, k_2, ..., k_m\}$ using a light-weight LLM to enhance retrieval effectiveness for complex questions referencing multiple schema elements, with prompt details provided in \ref{app:keyword}. These extracted keywords serve as independent retrieval queries that help capture relevant schema elements even when the full question's semantic representation doesn't closely match corresponding schema elements.

For each $e \in E$ we concurrently perform retrieval across each $k \in K \cup \{q\}$, with an ablation study in \ref{method:component_ablation} analyzing the impact of keyword-level and question-level retrieval. We also evaluate appending $k$ directly to $q$ as done in CHESS, but we find that this under-performs direct keyword-level retrieval.

\subsubsection{Entity-Type Relevance Calibration}
We hypothesize that each $\lambda \in \Lambda$ will have varying levels of importance within each dataset, which may not be captured by direct relevance scores. To account for this, we propose an entity-type relevance score calibration, where weights are calibrated over ground truth training samples when available. For each entity type $\lambda \in \Lambda$, we compute:

\begin{equation} w_\lambda = \frac{|\Lambda| \cdot \text{AUC}(\lambda)^2}{\sum_{\lambda' \in \Lambda} \text{AUC}(\lambda')^2} \end{equation}

where $\text{AUC}(\lambda)$ is the area under the table-level recall curve for entity type $\lambda$ over training data. We square the AUC values to amplify differences between entity types, giving greater weight to those with stronger predictive power. These weights are then applied to scale relevance scores at inference time, ensuring that entity types with consistently stronger predictive power receive higher influence in the final ranking.

Inspired by CRUSH \cite{crush}, we also explored keyword-level entropy-guided relevance calibration prior to entity-level calibration, which is designed to address the variable discriminative power of different keywords across schema entities. However, we found that this component did not improve system performance and have excluded it from our results; details on the component methodology and observed impact are discussed in \ref{app:entropy}.

\subsection{SQL Generation}

\textbf{Table Prediction}. While the resulting schema entities can be used directly to construct a schema for SQL generation, we find it is beneficial to perform an intermediary table prediction prior to final query generation.  This step constructs a candidate schema and applies an LLM to predict rank-ordered tables relevant to $q$, with specific prompts used detailed in \ref{app:table_pred}. Full schemas of tables identified as revelant are then loaded for final SQL generation. We find that this step is especially beneficial for covering unknown join relations which cannot be inferred from $q$, as well as better leveraging semantic context in $E_{\Lambda_T}$.

\textbf{SQL Generation}. Following retrieval, any SQL generation pipeline can be applied to the final schema, with our specific evaluations using zero-shot text-to-SQL with self-correction. Details on the specific prompt used in experiments is provided in \ref{app:sql_gen_ee}.

\section{Experiments}

\subsection{Dataset Details}

We evaluate our method over three benchmarks with dataset statistics provided in Table \ref{tab:dataset_stats}. The Spider and BIRD benchmarks are designed for the single-database setting, which we adapt to the massive catalog setting by considering the full set of training and test schema for each test record.

\begin{itemize} 
\item \textbf{Spider} \cite{spider}: A widely-used cross domain text-to-SQL benchmark with $\Lambda_C = \{\textit{column name}, \textit{column alias}\}, \: \Lambda_T = \\ \{\textit{table name}, \textit{table alias}\}$. Database schemas are well-named, with schema elements that are well-aligned with questions.

\item \textbf{BIRD} \cite{bird}: A challenging text-to-SQL benchmark emphasizing database content understanding. BIRD contains richer schema context designed to test text-to-SQL systems' abilities to incorporate domain knowledge, with $\Lambda_C = $ \{\textit{column name}, \textit{column alias}, \textit{column descr.}, \textit{value descr.}\}, \: $\Lambda_T = $ \{\textit{table name}\}. Questions may involve references to associated schema context, as opposed to direct column and table names as in Spider.
\item \textbf{Fiben} \cite{fiben}: An enterprise-focused benchmark developed by IBM across financial schemas. Fiben uses minimal metadata with $\Lambda_C = \{\textit{column name}\}, \: \Lambda_T = \{\textit{table name}\}$. Questions often lack context, making schema identification challenging.
\end{itemize}

For all datasets, we additionally synthesize $E_{\lambda_{\text{table descr.}}}$ from table schemas following the process in \ref{app:table_synth} and analyze the impact of adding it to $\Lambda_T$ on token consumption and performance in ablations.  Primary evaluations are reported with $\lambda_{\text{table descr.}}$ excluded for fair comparison.

\begin{table}[h]
\centering
\begin{tabular}{l|cc|cccc}
\hline
\multirow{2}{*}{Dataset} & \multicolumn{2}{c|}{Size} & \multicolumn{4}{c}{Schema} \\
\cline{2-7}
 & Train & Test & \shortstack{N\\DBs} & \shortstack{N\\Tables} & \shortstack{N\\Cols} & \shortstack{Avg Cols\\per Table} \\
\hline
Spider & 7000 & 1034\textsuperscript{a} & 166 & 876 & 4503 & 5.1 \\
BIRD & 9427 & 1534\textsuperscript{a} & 80 & 597 & 4337 & 7.26 \\
Fiben & 0 & 279 & 1 & 152 & 374 & 2.46 \\
\hline
\end{tabular}
\caption{Statistics of datasets. \: \textsuperscript{a} We use the original development sets for testing.}
\label{tab:dataset_stats}
\end{table}

\subsection{Baseline Methods}

We compare RASL to various retrieval baselines, including both general methods adapted to text-to-SQL and methods proposed specifically for text-to-SQL schema retrieval.
\begin{itemize}
    \item \textbf{BM25} \cite{bm25}: BM25 is a lexical retrieval method widely used for document retrieval and ranking. We adapt this to table retrieval by treating each table's complete schema as one document.
    \item \textbf{Sentence transformer (SXFMR)} \cite{sxfmr}: The sentence transformer is a semantic retrieval model designed to embed text to the same latent representation. Similar to BM25, table schemas are treated as documents for embedding.
    \item \textbf{CRUSH} \cite{crush}: CRUSH uses an LLM to hallucinate a schema from $q$, followed by retrieving target schema columns by lexical or semantic similarity to hallucinated schema columns. For table-level retrieval evaluations, we take the distinct tables corresponding to top ranking columns.
    \item \textbf{DTR} \cite{dtr}: DTR uses contrastive learning on ($q$, $t$) pairs to train a table retriever. This method requires synthesizing extensive training data over the target database, and may not be best suited for enterprise settings with continuously evolving databases.
    \item \textbf{DBCopilot} \cite{dbcopilot}: DBCopilot synthesizes text-to-SQL pairs over a hierarchical graph of known database and table relations and uses these to train a constrained decoder for table prediction. Similar to DTR, this requires synthesis of training data over the target database schema.
\end{itemize}

\subsection{Evaluation Details}

Consistent with previous works \cite{dbcopilot, crush, dtr}, we primarily evaluate our method using macro-average Recall@$N$ with respect to ground truth tables used in each SQL query, which measures the fraction of relevant instances in the top-$N$ predicted tables.  For primary evaluations on Recall@$N$, we directly adopt the metrics reported in \cite{dbcopilot}, where specific method configurations are provided in \ref{app:baseline_details} and RASL is applied over identical testing sets. For ablation studies, we focus on comparison to retrieval-based baselines which do not involve model fine-tuning, as these methods are most relevant to industry settings with evolving catalog schemas. For all ablations, we closely follow the implementation details reported in \cite{dbcopilot}, with the exception of using Anthropic Claude 3.5 Sonnet-v2 \cite{claude_sonnet} instead of OpenAI GPT-3.5-turbo \cite{openai} for CRUSH schema hallucination due to model access constraints.

\begin{table}[h]
\centering
\small
\begin{tabular}{l|cc|cc|cc}
\hline
\multirow{2}{*}{Model} & 
\multicolumn{2}{c|}{Spider} & 
\multicolumn{2}{c|}{BIRD} & 
\multicolumn{2}{c}{Fiben} \\
\cline{2-7}
 & R@5 & R@15 & R@5 & R@15 & R@5 & R@15 \\

\hline
BM25 & 
86.5 & 93.9 & 
68.3 & 82.8 & 
33.3 & 38.6 \\

SXFMR & 
80.4 & 92.3 & 
67.6 & 83.1 & 
28.2 & 46.5 \\

\hline\rule{0pt}{2.5ex}

$\text{CRUSH}_{\text{BM25}}$ & 
87.2 & 95.0 & 
68.4 & 87.8 &
34.9 & 54.0 \\

$\text{CRUSH}_{\text{SXFMR}}$ & 
82.2 & 93.9 & 
70.6 & 85.1 & 
34.1 & 50.8 \\

\hline\rule{0pt}{2.5ex}

DTR & 
76.3 & 93.2 & 
76.2 & 92.0 & 
37.7 & 48.9 \\

DBCopilot & 
91.6 & 97.6 & 
85.8 & 94.6 & 
41.1 & 56.9 \\

\hline\rule{0pt}{2.5ex}

$\text{RASL}_{\text{retriever}}$ & 
72.1 & 94.1 & 
70.5 & 92.6 & 
34.6 & 64.5 \\

$\text{RASL}_{\text{full}}$ & 
\textbf{97.0} & \textbf{98.0} & 
\textbf{97.5} & \textbf{97.8} & 
\textbf{69.1} & \textbf{69.2} \\
\hline
\end{tabular}
\caption{Primary performance comparison measuring Recall@$N$, where $N$ is the maximum number of tables. RASL is applied without $\lambda_{\text{table descr.}}$ in this setting.}
\label{tab:overall_results}
\end{table}

\subsection{Experiment Settings}

For all experiments, RASL uses Anthropic Claude 3.5 Haiku \cite{claude_haiku} for keyword extraction, Cohere Embed-v3-English \cite{cohere} in OpenSearch Serverless through Amazon Bedrock Knowledge Bases \cite{bedrock_kb} for vector database embedding, and Anthropic Claude 3.5 Sonnet-v2 \cite{claude_sonnet} for table prediction and SQL generation; temperature is set to 0.0 throughout for reproducibility.

For all datasets except Fiben, which does not contain a training set, we perform entity-type weight calibration using 200 randomly sampled training instances. During retrieval, the top 100 entities within each keyword and entity type are retrieved prior to relevance calibration and filtering, as this is the maximum allowed by Bedrock Knowledge Bases.

\subsection{Table Retrieval Results}

\begin{figure*}[h]
    \centering
    \includegraphics[width=1.0\textwidth]{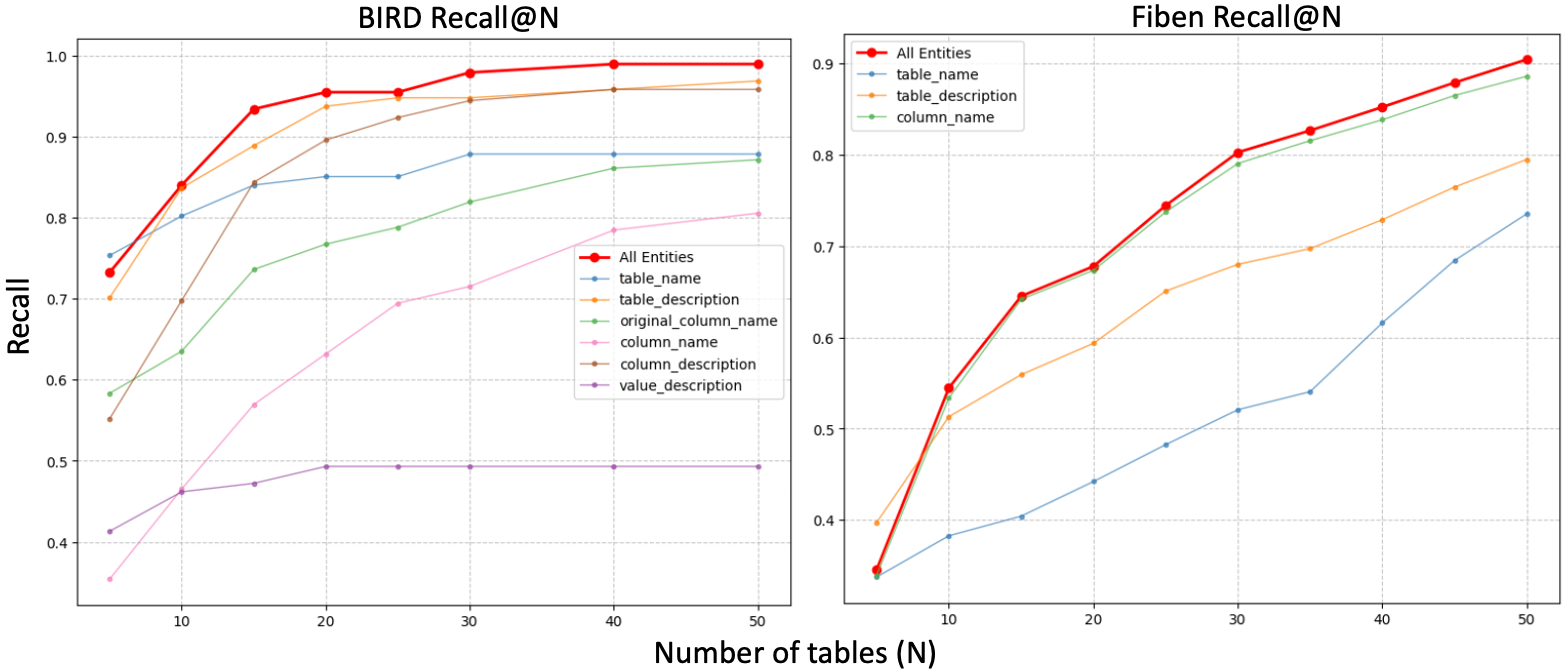}
    \caption{Table $Recall@N$ over BIRD (left) and Fiben (right). BIRD benefits from both $\Lambda_C$ and $\Lambda_T$, achieving notable recall improvement over individual $\lambda$ at higher $N$, while Fiben primarily leverages $E_C$.}
    \label{fig:recall_at_n}
\end{figure*}

In Table \ref{tab:overall_results} we evaluate RASL against baseline systems for table retrieval. We evaluate RASL in two settings: (1) \textit{retriever-only}, which ranks tables by calibrated relevance scores across all entities $E$, and (2) \textit{full-system}, which applies table prediction after filtering $E$ to entities from the top $N$ ranked tables. For \textit{full-system}, we limit the retrieved candidate entities to $N=50$ tables, as we find that this consistently keeps schema entities below $3\%$ per type (\ref{app:entity_usage}) while maintaining high table recall, ensuring manageable prompt lengths for prediction.

$\text{RASL}_\text{full}$ out-performs all baselines, with similar $R@5$ and $R@15$, demonstrating the effectiveness of our dual-stage retrieve-then-predict apporach for precise table identification using rich granular context. While $\text{RASL}_\text{retriever}$ shows lower recall at $N=5$, analyses in \ref{app:conflict_analysis} attributes this to highly overlapping columns/tables across databases, with performance improving at higher $N$ and consistently ranking among the top three retrieval-based methods at $N=15$. This stronger recall at higher $N$ proves particularly valuable when combined with entity-level schema construction, as we can significantly reduce context while preserving the most informative schema entities for table prediction and SQL generation. Figure \ref{fig:recall_at_n} further demonstrates the advantages of our combined entity retrieval approach, which outperforms entity-specific methods on both context-rich (BIRD) and context-sparse (Fiben) datasets.

\subsection{Impact of Retrieval Mechanism on Table Prediction}

\begin{table}[h]
\centering
\small
\begin{tabular}{l|ccc|ccc}
\hline
\multicolumn{7}{c}{Without RASL Table Descriptions} \\
\hline
\multirow{2}{*}{Model} & 
\multicolumn{3}{c|}{LLM-Filtered Prediction} &
\multicolumn{3}{c}{Initial Retrieval Pool} \\
\cline{2-7}
 & Spider & BIRD & Fiben & Spider & BIRD & Fiben \\
 & (R@5) & (R@5) & (R@5) & (R@N) & (R@N) & (R@N) \\
\hline
BM25 & 87.0 & 71.8 & 41.5 & 87.7 & 72.3 & 43.4 \\
SXFMR & 88.0 & 66.9 & 40.1 & 89.0 & 70.1 & 41.8 \\
$\text{CRUSH}_{\text{BM25}}$ & 91.7 & 83.5 & 57.1 & 96.4 & 86.7 & 61.1 \\
$\text{CRUSH}_{\text{SXFMR}}$ & 94.2 & 93.4 & 58.2 & 97.6 & 97.7 & 73.7 \\
$\text{RASL}_{\text{full}}$ & \textbf{97.0} & \textbf{97.5} & \textbf{69.1} & 99.3 & 98.1 & 90.6 \\

\hline
\multicolumn{7}{c}{With RASL Table Descriptions} \\
\hline
\multirow{2}{*}{Model} & 
\multicolumn{3}{c|}{LLM-Filtered Prediction} &
\multicolumn{3}{c}{Initial Retrieval Pool} \\
\cline{2-7}
 & Spider & BIRD & Fiben & Spider & BIRD & Fiben \\
 & (R@5) & (R@5) & (R@5) & (R@N) & (R@N) & (R@N) \\
\hline
BM25 & 95.6 & 84.8 & 68.9 & 96.7 & 86.5 & 99.9 \\
SXFMR & 96.2 & 83.2 & 70.0 & 99.1 & 87.8 & 99.9 \\
$\text{CRUSH}_{\text{BM25}}$ & 95.4 & 87.8 & 70.7 & 100.0 & 95.0 & 99.1 \\
$\text{CRUSH}_{\text{SXFMR}}$ & 94.5 & 93.1 & 67.1 & 100.0 & 99.5 & 99.9 \\
$\text{RASL}_{\text{full}}$ & \textbf{98.2} & \textbf{97.5} & \textbf{77.6} & 99.1 & 98.9 & 90.4 \\
\hline
\end{tabular}
\caption{Performance comparison across datasets with and without RASL table descriptions. LLM-Filtered Prediction shows the recall@5 after filtering, while Initial Retrieval Pool shows the table recall@N of the initial retriever output prior to filtering, where N varies by model and dataset.}
\label{tab:context_budget_ablation}
\end{table}

We believe that RASL's primary strength lies in efficiently loading relevant context for precise table identification within manageable context budgets, rather than standalone context retrieval. To validate this, we compare $\text{RASL}_\text{retriever}$ against BM25, SXFMR, and CRUSH for context retrieval prior to LLM-based table prediction. However, since these methods operate at different granularities--RASL employs multi-entity retrieval, BM25 and SXFMR work at the table level, and CRUSH operates at the column level--we implement a standardized protocol to ensure fair comparison across these diverse approaches:
\begin{itemize}
\item For RASL: Filter entities to those from top N=50 tables by relevance
\item For baselines: Add schema elements until reaching RASL's context budget
    \begin{itemize}
        \item BM25/SXFMR: Add full table schemas
        \item CRUSH: Add column names, where the first column includes the table name
        \item All baselines: Include all benchmark-provided $E_\Lambda$ in constructed schemas
    \end{itemize}
\end{itemize}

Given $\lambda_{\text{table descr.}}$'s substantial length compared to other entity types, we evaluate performance both with and without it, adjusting baseline context budgets accordingly. While this approach doesn't guarantee identical contexts, it ensures RASL's context size never exceeds baselines.

Results in Table \ref{tab:context_budget_ablation} show RASL significantly outperforming baselines in both settings, demonstrating strong synergy between multi-entity retrieval and table prediction. Without table descriptions, RASL achieves higher initial retrieval recall than all baselines under equal context budgets. With table descriptions included, baselines often achieve higher initial recall by fitting more distinct schemas within the budget. However, RASL still achieves superior table prediction performance, suggesting it retrieves more relevant context. This is further evidenced by RASL's larger improvements in prediction recall versus initial pool recall when including table descriptions, indicating these descriptions provide unique value beyond other retrieved entities.

\subsubsection{Impact of Table Descriptions on Context Usage}
\label{method:table_descr_ablation}

While including $\lambda_{\text{table descr.}}$ improves RASL's table prediction performance (Table \ref{tab:context_budget_ablation}), it significantly increases token consumption. Analysis of token usage statistics in Table \ref{tab:token_usage} reveals that consumption patterns strongly depend on dataset structure, and given RASL's fixed N=50 table retrieval limit, $\lambda_{\text{table decr.}}$ token usage inversely correlates with average columns per table. For BIRD, which has 7.26 columns/table, table descriptions consume only 3\% of the original schema tokens. In contrast, for Fiben, with just 2.46 columns/table and significantly smaller total database size, table descriptions exceed the entire original schema size.

While experiments validate the utility of comprehensive table descriptions, their token consumption generally outweighs performance benefits across tested datasets. Though this approach may prove valuable for settings with wide tables or high-accuracy requirements, we believe future work on more concise table context synthesis could better serve RASL under tight context budgets.

\begin{table}[h]
\centering
\resizebox{\columnwidth}{!}{
\begin{tabular}{l|cc|cc|cc}
\hline
& \multicolumn{2}{c|}{Spider} & \multicolumn{2}{c|}{BIRD} & \multicolumn{2}{c}{Fiben} \\
\cline{2-7}
\multirow{2}{*}{Component} & 
Toks. & Avg. Toks. & Toks. & Avg Toks. & Toks. & Avg Toks. \\
& (Total) & (N=50) & (Total) & (N=50) & (Total) & (N=50) \\
\hline
Table name & 3,120 & 63 & 1,594 & 39 & 1,180 & 210 \\
Table alias & 3,178 & 44 & - & - & - & - \\
Column name & 15,143 & 65 & 11,257 & 50 & 1,887 & 339 \\
Column alias & 15,657 & 93 & 10,105 & 38 & - & - \\
Column descr. & - & - & 34,903 & 202 & - & - \\
Value descr. & - & - & 28,379 & 55 & - & - \\
Table descr. & 259,885 & 3,670 & 146,442 & 2,357 & 37,780 & 6,441 \\
\hline
Total without & & & & & & \\
table descr. & 37,098 & 266 & 86,238 & 384 & 3,067 & 549 \\
\hline
Total with & & & & & & \\
table descr. & 296,983 & 3,936 & 232,680 & 2,741 & 40,847 & 6,990 \\
\hline
\end{tabular}
}
\caption{Token usage breakdown by schema components with and without including $\lambda_{\text{table descr.}}$. We compare the total database schema tokens to the average tokens used by $\text{RASL}_{\text{retriever}}$ at $N=50$. Each token is approximately 3.5 characters.}
\label{tab:token_usage}
\end{table}

\subsection{RASL Component Ablation}
\label{method:component_ablation}

Table \ref{tab:component_ablation} presents our analysis of keyword-level retrieval and entity-type relevance score calibration. Maintaining our experimental setup of $\text{RASL}_{\text{full}}$ with N=50 table filtering, we evaluate both final table prediction and initial retrieval pool recall. Results show keyword-based retrieval ($K$) significantly outperforms question-based retrieval ($q$) across all datasets, demonstrating the importance of granular search queries for $E$ retrieval. While combining both approaches ($K \cup \{q\}$) yields slight improvements in most settings, keyword-based retrieval remains the primary driver of performance.

The impact of entity-type weight calibration ($W_{\Lambda}$) varies by dataset, providing substantial gains for Spider but only modest improvements for BIRD; results are excluded for Fiben due to no training samples being available. We observe that while $W_{\Lambda}$ can be beneficial, it is not a necessary component of RASL, which demonstrates strong utility even in settings where no labeled data exists.

\begin{table}[h]
\centering
\small
\begin{tabular}{ll|ccc}
\hline
& & \multicolumn{2}{c}{$\text{RASL}_{\text{full}}$} & $\text{RASL}_{\text{retriever}}$ \\
Dataset & Configuration & R@5 & R@15 & R@50 \\
\hline
\multirow{4}{*}{Spider} 
& $K \cup \{q\}$ & \textbf{97.0} & 98.0 & \textbf{99.3} \\
& $K$ only & 96.6 & \textbf{98.2} & 99.1 \\
& $q$ only & 92.7 & 93.2 & 97.8 \\
& $K \cup \{q\}$ w/o $W_\Lambda$ & 94.3 & 95.8 & 96.8 \\
\hline
\multirow{4}{*}{BIRD}
& $K \cup \{q\}$ & \textbf{97.5} & \textbf{97.8} & 98.1 \\
& $K$ only & 96.8 & 97.1 & \textbf{98.3} \\
& $q$ only & 90.1 & 90.1 & 95.8 \\
& $K \cup \{q\}$ w/o $W_\Lambda$ & 97.2 & 97.4 & 98.0 \\
\hline
\multirow{4}{*}{Fiben}
& $K \cup \{q\}$ & \textbf{69.1} & \textbf{69.2} & \textbf{90.6} \\
& $K$ only & \textbf{69.1} & \textbf{69.2} & 89.7 \\
& $q$ only & 67.4 & 67.5 & 79.4 \\
& $K \cup \{q\}$ w/o $W_\Lambda$ & - & - & - \\
\hline
\end{tabular}
\caption{Ablation study on the impact of retrieval query type ($K$ vs. $q$) and entity-level weight calibration ($W_{\Lambda}$). $\text{RASL}_{\text{full}}$ filters $E$ to the top $N=50$ tables prior to table prediction, with recall reported over final table prediction and initial candidate tables.}
\label{tab:component_ablation}
\end{table}

\subsection{Error Analysis}

Next we cover some common error cases observed across baseline methods. In Figure \ref{fig:error1} we show how assumptions in CRUSH schema hallucination can impact retrieval performance. In this case, we see that an incorrect \textit{person} table name causes all segments to over-index on people-related columns and associated tables. In contrast, RASL uses granular and isolated keywords directly extracted from the question. When paired with entity-level retrieval, this allows for highly relevant specific tables and columns to be loaded from any keywords extracted from the question, without assumptions on how the schema is structured.

In Figure \ref{fig:error2} and \ref{fig:error3} we show the most common causes of error in table retrieval baselines, which is insufficient granular retrieval context. We see that important keywords, such as \textit{circuits} in Figure \ref{fig:error2} or \textit{cards} in Figure \ref{fig:error3} are not sufficiently captured in table-level similarities, resulting in necessary tables being missed.

\begin{tcolorbox}[
colback=gray!10,
colframe=black,
title=Incorrect CRUSH hallucinated schema,
after=\captionof{figure}{Example of schema hallucination leading to poor matching by CRUSH.}\label{fig:error1}
]
\small
\raggedright
\textbf{\textit{Question}}: Where is Amy Firth's hometown? Hometown refers to city, county, state \\
\vspace{5pt}
\textbf{\textit{{Ground Truth Tables}}}: ['student\_club.member', 'student\_club.zip\_code'] \\
\vspace{5pt}
\textbf{\textit{{Ground Truth Columns}}}: ['member.first\_name', 'member.last\_name', 'member.zip', 'zip\_code.city', 'zip\_code.state', 'zip\_code.county', 'zip\_code.zip\_code'] \\
\vspace{5pt}
\hrule
\vspace{5pt}
\textbf{$\text{CRUSH}_{\text{SXFMR}}$} \\
\vspace{5pt}
\textbf{\textit{Hallucinated Schema}}: ['person.first\_name', 'person.last\_name', 'person.hometown\_city', 'person.hometown\_county', 'person.hometown\_state'] \\
\vspace{5pt}
\textbf{\textit{Top 5 Tables}}: ['human\_resources.employee',
 'works\_cycles.person',
 'movie\_3.actor',
 'address.state',
 'address.country'] \\
\textbf{\textit{Index of Correct Table}}: \{'student\_club.member': 31, 'student\_club.zip\_code': 30\} 
\vspace{5pt}
\hrule 
\vspace{5pt}
\textbf{RASL} \\
\vspace{5pt}
\textbf{\textit{Keywords}}: ['person' 'town', 'city', 'county', 'state', 'name', 'first name', 'last name', 'location'] \\
\vspace{5pt}
\textbf{\textit{Top 5 Tables}}: ['law\_episode.person',
 'regional\_sales.`store locations`',
 \textbf{'student\_club.zip\_code'},
 \textbf{'student\_club.member'},
 'retail\_complains.district']
\end{tcolorbox}

\begin{tcolorbox}[
colback=gray!10,
colframe=black, 
title=Failure of Semantic Table Retrieval,
after=\captionof{figure}{Failure of semantic table retriever due to lack of granular context.}\label{fig:error2}
]
\small
\raggedright
\textbf{\textit{Question}}: How many formula\_1 races took place on the circuits in Italy?
 \\
 \vspace{5pt}
\textbf{\textit{{Ground Truth Tables}}}: ['formula\_1.circuits', 'formula\_1.races'] \\
\vspace{5pt}
\textbf{\textit{{Ground Truth Columns}}}: ['formula\_1.circuits.circuitid', 'formula\_1.circuits.country', 'formula\_1.races.circuitid'] \\
\vspace{5pt}

\hrule
\vspace{5pt}
\textbf{SXFRMR} \\
\vspace{5pt}
\textbf{\textit{Top 3 Tables}}: ['formula\_1.qualifying', 'formula\_1.results',
       'formula\_1.constructorstandings'] \\
\vspace{5pt}
\textbf{\textit{Index of Correct Table}}: \{'formula\_1.races': 4, 'formula\_1.circuits': 24\} 
\vspace{5pt}
\hrule 
\vspace{5pt}
\textbf{RASL} \\
\vspace{5pt}
\textbf{\textit{Keywords}}: ['formula\_1', 'races', 'country', 'circuits', 'race circuits'] \\
\vspace{5pt}
\textbf{\textit{Top 3 Tables}}: ['formula\_1.circuits', 'formula\_1.races', 'formula\_1.qualifying'] \\
\end{tcolorbox}

\begin{tcolorbox}[
colback=gray!10, 
colframe=black, 
title=Failure of Lexical Retrieval,
after=
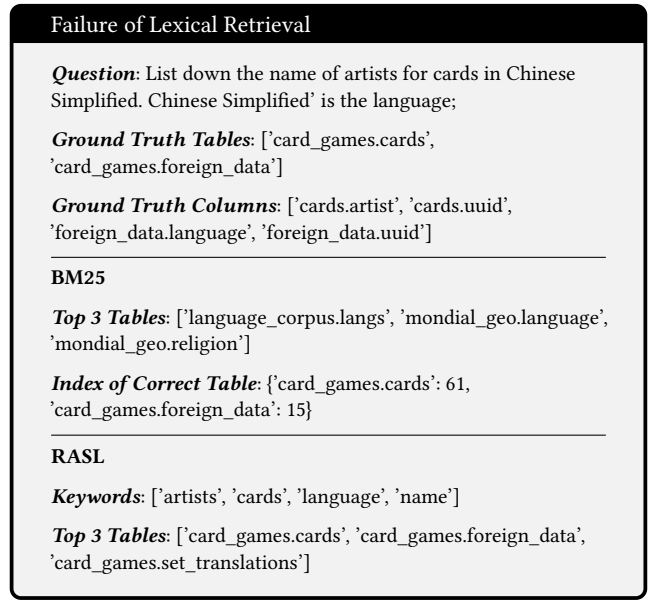
\captionof{figure}{Failure of lexical table retriever due to lack of granular context.}\label{fig:error3}
]
\small
\raggedright
\textbf{\textit{Question}}: List down the name of artists for cards in Chinese Simplified. Chinese Simplified' is the language;
 \\
 \vspace{5pt}
\textbf{\textit{{Ground Truth Tables}}}: ['card\_games.cards', 'card\_games.foreign\_data'] \\
\vspace{5pt}
\textbf{\textit{{Ground Truth Columns}}}: ['cards.artist', 'cards.uuid', 'foreign\_data.language', 'foreign\_data.uuid'] \\
\vspace{5pt}
\hrule
\vspace{5pt}
\textbf{BM25} \\
\vspace{5pt}
\textbf{\textit{Top 3 Tables}}: ['language\_corpus.langs', 'mondial\_geo.language', 'mondial\_geo.religion'] \\
\vspace{5pt}
\textbf{\textit{Index of Correct Table}}: \{'card\_games.cards': 61, 'card\_games.foreign\_data': 15\}

\vspace{5pt}
\hrule 
\vspace{5pt}
\textbf{RASL} \\
\vspace{5pt}
\textbf{\textit{Keywords}}: ['artists', 'cards', 'language', 'name'] \\
\vspace{5pt}
\textbf{\textit{Top 3 Tables}}: ['card\_games.cards', 'card\_games.foreign\_data', 'card\_games.set\_translations'] \\
\end{tcolorbox}

\subsection{End-to-End SQL Generation}

In Table \ref{tab:sql_gen_results} we compare the performance of RASL to retrieval-based baselines on end-to-end SQL generation. This evaluation applies the full RASL pipeline with the same setting as previously, where all $E$ belonging to the top $N=50$ tables by relevance are used for table prediction. For SQL generation we load the full original schema for each predicted table, retaining only $\Lambda$ provided by the benchmarks and excluding $\lambda_{\text{table descr.}}$ from RASL due to increased token consumption outweighing benefits for practical enterprise applications. Due to relatively low recall at $N=5$ for all baseline methods, we evaluate performance over $N=15$ and $N=30$, as well as compare to the standard single database text-to-SQL setting where all tables corresponding to the target database are loaded. We report text-to-SQL execution accuracy and table recall with respect to ground truth SQL queries as our primary metrics. We also list the average number of tokens used to construct schemas used in prompting. Since RASL contains two prompting steps (table prediction and SQL generation), we sum the total schema tokens over both steps for RASL.

\begin{table*}[h]
\centering
\small
\begin{tabular}{l|ccc|ccc|ccc|ccc}
\hline
& \multicolumn{6}{c|}{BIRD} & \multicolumn{6}{c}{Spider} \\
\cline{2-13}
& \multicolumn{3}{c}{N=15} & \multicolumn{3}{c|}{N=30} & \multicolumn{3}{c}{N=15} & \multicolumn{3}{c}{N=30} \\
\cline{2-13}
Model & Acc. & Recall & Tokens & Acc. & Recall & Tokens & Acc. & Recall & Tokens & Acc. & Recall & Tokens \\
\hline
BM25 & 47.4 & 84.6 & \textbf{4,057} & 49.5 & 89.3 & 8,577 & 58.3 & 87.0 & 1,304 & 59.3 & 89.2 & 2,662 \\
SXFMR & 43.6 & 82.6 & 4,989 & 48.0 & 89.1 & 8,993 & 58.0 & 85.7 & \textbf{1,182} & 59.0 & 87.9 & 2,364 \\
$\text{CRUSH}_{\text{BM25}}$ & 43.5 & 80.92 & 5,133 & 50.1 & 87.5 & 9,978 & 60.1 & 89.3 & 1,395 & 62.5 & 90.3 & 2,764 \\
$\text{CRUSH}_{\text{SXFMR}}$ & 49.4 & 91.6 & 5,589 & 52.9 & 93.0 & 10,385 & 55.9 & 87.0 & 1,374 & 58.0 & 91.6 & 2,744 \\
$\text{RASL}_{\text{full}}$ & \textbf{53.5} & \textbf{94.6} & 4,696 & \textbf{53.5} & \textbf{94.6} & \textbf{4,696} & \textbf{64.5} & \textbf{92.5} & 1,266 & \textbf{64.5} & \textbf{92.5} & \textbf{1,266} \\
\hline
Gold & 54.0 & 96.7 & 2,489 & 54.0 & 96.7 & 2,489 & 69.0 & 99.5 & 389 & 69.0 & 99.5 & 389 \\
\hline
\end{tabular}
\caption{Execution accuracy and table recall of RASL versus baselines on end-to-end SQL generation. Token counts represent total schema tokens consumed for each method, including both table prediction and SQL generation steps for RASL. Bold indicates best method.}
\label{tab:sql_gen_results}
\end{table*}

We see that RASL consistently ranks best in both SQL table recall and execution accuracy compared with all baselines. We note that for both BIRD and Spider, RASL never predicts over 15 tables, accounting for identical metrics for both $N=15$ and $N=30$. We also observe that RASL constructs token-efficient schemas, being the most efficient method at $N=30$ and ranking second at $N=15$. For both Spider and BIRD, we see that on average the table prediction schema accounts for approximately $2/3$ of the total schema tokens, with the final SQL generation over full schemas from selected tables accounting for the other $1/3$. A detailed cost analysis based on commercial API pricing is provided in Appendix~\ref{app:cost_analysis}.


\section{Discussion and Next Steps}

RASL successfully addresses the challenge of scaling text-to-SQL systems to enterprise-level databases through an effective component-based retrieval architecture. Our experiments demonstrate significant performance improvements over baseline retrieval-based methods while maintaining practical context budgets. We outline key insights, limitations, and future research directions that emerge from our work.

\subsection{Key Insights}

\textbf{Entity-level decomposition} proves highly effective for context retrieval under token constraints. By decomposing both questions and schemas into granular semantic units, RASL preserves critical information while scaling to massive schemas without requiring domain-specific training. This approach shows particular strength in context-rich environments like BIRD, where entity-level retrieval achieves greater recall improvements over table-level methods, which often fail at covering important details within user questions.

\textbf{Two-stage retrieval-prediction} creates a powerful synergy that consistently outperforms alternative methods. Our approach first narrows the search space through multi-entity retrieval, then applies LLM reasoning to identify the most relevant tables. While $\text{RASL}_{\text{retriever}}$ does not excel directly at stand-alone high-precision table identification, we find that it retrieves more valuable context than baseline retrieval methods under the same token budget, with improved performance on table prediction. Furthermore, we show that our end-to-end system consistently enables higher SQL generation accuracy than baseline methods while consuming fewer overall input tokens.

\subsection{Limitations}

While we validate that synthesizing additional semantic context such as table descriptions can further improve performance, their inclusion creates significant token consumption trade-offs, particularly for databases with few columns per table. We believe that although promising, the proposed approach requires more refinement before additional synthesized context can provide low-cost performance benefits.

For all evaluations, we apply RASL to retrieve context relating to the top $N=50$ tables by relevance. While absolute token usage at this setting is significantly lower than LLM context window budgets, we believe that a deeper analysis of performance over different $N$ values would help identify optimal trade-offs between context size and retrieval effectiveness.

Lastly, while RASL proves effective using independent entity retrieval, we believe that further information may be contained in relative relevance across entity types. For instance, it may be beneficial to increase column-level entity scores when multiple distinct entities are retrieved from the same table.

\subsection{Future Directions}

Several promising research directions emerge from our work. Synthesizing more concise and granular table context entities may improve retrieval quality while keeping token consumption manageable. Exploring dynamic entity-level token allocation based on database characteristics could further enhance performance. Additional opportunities include extending evaluation of RASL over cross-database queries, integrating with recent advances in SQL generation techniques (e.g., multi-prompting and self-verification), and using agentic retrieval approaches to iteratively retrieve context using guided keyword searches based on past retrieval observations.

RASL represents a significant step toward practical natural language interfaces for massive database environments. By addressing the critical bottleneck of schema linking at scale without requiring specialized fine-tuning, it enables more accessible deployment of text-to-SQL systems across diverse enterprise settings. Future work building on this foundation has the potential to further bridge the gap between natural language understanding and database access.


\section{Conclusion}

In this work, we present RASL, a zero-shot framework for scaling natural language querying to massive databases. RASL decomposes database schemas into granular semantic entities, retrieves relevant context via calibration-enhanced similarity to important question components, and reasons over the resulting reduced schema for predicting relevant tables and generating SQL queries. We demonstrate that RASL out-performs baselines across multiple datasets varying in database size, relational information, and available semantic context. While prior works addressing this challenge often rely on domain-specific fine-tuning, which complicates deployment, RASL is designed to be robust to database schema and context changes--only requiring syncing the vector database with no model training required--and can be easily deployed in serverless computing environments.

\bibliography{references}

\clearpage

\appendix

\section{Schema Entity Usage by $N=50$ Top Relevant Tables}
\label{app:entity_usage}

Below we show the percent of total database schema entities used with entities filtered by top $N=50$ distinct tables over Spider and BIRD benchmarks. We see that total schema is consistently reduced to less then 3\% of the total schema entities for $\Lambda_T$, and less than 1\% of total schema entities for $\Lambda_C$.

\begin{table}[h]
\centering
\small
\begin{tabular}{l|cc}
\hline
& \multicolumn{2}{c}{Dataset} \\
\cline{2-3}
Entity Type & Spider & BIRD  \\
\hline
Table Name & 1.78\% & 1.99\%  \\
Table Alias & 1.27\% & -  \\
Table Descr. & 1.32\% & 2.18\%  \\
Column Name & 0.50\% & 0.42\%  \\
Column Alias & 0.66\% & 0.37\%  \\
Column Descr. & - & 0.49\%  \\
Value Descr. & - & 0.29\%  \\
\hline
\end{tabular}
\caption{Entity usage across different datasets}
\label{tab:entity_usage}
\end{table}

\section{Exploration of Entropy-Guided Keyword-Level Weight Calibration}
\label{app:entropy}

Information retrieval systems often struggle with keywords that have varying levels of specificity--some terms may be highly specific to certain schema elements, while others may be generic and match well with many elements. To address this challenge, we explored applying the entropy-guided similarity approach introduced by CRUSH \cite{crush}, adapting it to handle our keyword-level retrieval approach.

Building on CRUSH's core insight that keyword specificity should influence matching weights, we developed a calibration system that automatically identifies and adjusts for differences in keyword discriminative power. For example, a keyword like "latitude" that strongly matches only geographic coordinates should have more influence than "value" which may match well with many different schema elements.

We quantify this specificity using an information theory approach. For each keyword $k$ and entity type $\lambda$, we first compute a probability distribution over matching entities:

\begin{equation} p(e|k,\lambda) = \frac{\exp(\alpha \cdot r_0(e, k))}{\sum_{e' \in C_{k,\lambda}} \exp(\alpha \cdot r_0(e', k))} \end{equation}

where $r_0(e,k)$ is the initial relevance score and $\alpha$ is a scaling factor that helps differentiate between similar scores. We then measure how focused or diffuse these matches are using entropy:

\begin{equation} H(k,\lambda) = -\sum_{e \in C_{k,\lambda}} p(e|k,\lambda) \log p(e|k,\lambda) \end{equation}

A low entropy indicates the keyword strongly prefers certain schema elements (high specificity), while high entropy suggests the keyword matches broadly across many elements (low specificity).

We adjust the final relevance scores:

\begin{equation} r(e, k) = r_0(e, k) \cdot \sigma(\alpha(\bar{H}_{\lambda} - H(k,\lambda))) \end{equation}

where $\sigma$ is the sigmoid function and $\bar{H}_{\lambda}$ is the mean entropy for entity type $\lambda$. This calibration is designed to automatically reduce the influence of generic keywords while amplifying the impact of specific ones. The scaling parameter $\alpha$ controls the sharpness of the probability distribution.

While beneficial in CRUSH, we found that the inclusion of keyword-level entropy calibration had minimal impact on performance, with marginal impact at low $\alpha$ (1.0) and negative impact at higher $\alpha$ (2.0, 3.0, 5.0, 10.0). One possible explanation for this is that sharp distributions with few informative $e$ for a given $(k,\lambda)$ will result in all $e$ from that $(k,\lambda)$ being up-weighted, leading to more non-informative $e$ being included. Another possible explanation is that flat distributions with many high relevance $e$ may not necessarily indicate they are uninformative. For instance, we see in \ref{app:conflict_analysis} that there can be commonly named tables and columns across many databases. However, it may be more beneficial to include these at higher retrieval budgets, rather than down-weighting them due to lower specificity. For these reasons, we have excluded this component from our results, but we believe additional refinement of this method in future works may benefit RASL's performance.

\section{LLM Prompts}

\subsection{Table Prediction Prompt}
\label{app:table_pred}

\begin{tcolorbox}[
    breakable,
    colback=gray!10,
    colframe=gray!10,
    left=6pt,
    right=6pt,
    top=6pt,
    bottom=6pt
]
You are a database expert assistant that helps identify which tables are relevant to answering SQL questions. \\

TASK: \\
Analyze the provided database schema and the user's question, then identify which specific tables (and their databases) are most relevant for answering the question. You must rank tables in strict order of relevance. \\

\#\#\# Schema: \\
\{SCHEMA\} \\

\#\#\# User Question: \\
\{QUESTION\} \\

\#\#\# Instructions: \\
1. Examine the question carefully to understand what data would be needed to answer it \\
2. Analyze the database schema to determine which tables contain relevant information \\
3. Rank tables by relevance - tables listed first should be most central to answering the question \\
4. Consider both direct mentions and implied data needs \\
5. Select only tables that would contribute to a SQL query answering the question \\
6. Consider join paths needed to connect relevant information \\
7. IMPORTANT: The table schemas are incomplete and only contain possibly relevant columns. There are many columns not shown within each table. \\

Ranking Criteria: \\
- Primary tables: Directly contain data explicitly asked for in the question \\
- Secondary tables: Needed for joins or containing supplementary information \\ 
- Tertiary tables: Might be useful for contextual information but not essential \\

First, think through your reasoning step by step. Carefully consider how to rank the relevance of each table. \\

Then provide your answer in the following XML format: \\

<thinking> \\
Your detailed analysis explaining why specific tables are relevant to the question and how you determined their ranking order. \\
</thinking> \\

<relevant\_tables> \\
  <database name="database\_name"> \\
    <!-- Tables in strict order of relevance, most relevant first --> \\
    <table rank="1">most\_relevant\_table</table> \\
    <table rank="2">second\_most\_relevant\_table</table> \\
    <table rank="3">third\_most\_relevant\_table</table> \\
  </database> \\
  <database name="another\_database\_name"> \\
    <table rank="1">most\_relevant\_table\_in\_this\_db</table> \\
    <!-- Additional tables in decreasing relevance --> \\
  </database> \\
</relevant\_tables> \\

IMPORTANT: \\
- List tables in strict order of decreasing relevance within each database \\
- The "rank" attribute should reflect overall relevance across all databases (1 = most relevant overall) \\
- Only include tables that are genuinely relevant to answering the question \\
- If no tables from a particular database are relevant, do not include that database \\
- Only a subset of columns are shown for each table. Leverage these when relevant, but DO NOT assume the table is missing any columns
\end{tcolorbox}

\subsection{End-to-End SQL Generation Prompt}
\label{app:sql_gen_ee}

Below we show the prompt for end-to-end SQL generation. Since all datasets evaluated contain single-database target SQLs, we apply a prompt while involves first predicting the correct database, and then predicting the correct SQL over that database. For self-correction performed in experiments, we apply the same system prompt, with follow-up messages on the specific error encountered, or that the output table is empty if no results are returned, until the number of self-correction iterations is reached or a populated table is returned.

For extensions to cross-database querying using compatible SQL engines, the database prediction tag can be removed and instructions added on how to properly format tables with database name prefixes in the generated SQL query. There is no single-database constraint on context retrieval or schema construction.

\begin{tcolorbox}[
    breakable,
    colback=gray!10,
    colframe=gray!10,
    left=6pt,
    right=6pt,
    top=6pt,
    bottom=6pt
]
You are a data science expert. \\
Below, you are presented with a database schema and a question. \\
Your task is to generate a SQL query to answer the question. \\
\{DIALECT\_INSTRUCTION\} \\

\#\#\# Database Schema \\
\{DATABASE\_SCHEMA\} \\

\#\#\# Question \\
\{QUESTION\} \\

First, think through your reasoning step by step. If there are multiple databases, determine which one you should use, and then carefully consider how to rank the relevance of each table.  \\
Then provide your answer in the following XML format: \\

<thinking> \\
Your detailed analysis explaining why specific tables are relevant to the question and how you determined their ranking order. \\
</thinking> \\

<database> \\
The database you are executing the SQL query on \\
</database> \\

<sql\_query> \\
Your executable SQL query \\
</sql\_query> \\

IMPORTANT: \\
- Pay close attention to the specific columns used for selections and filtering, ensuring they are the correct ones. \\
- Pay close attention to any value formats provided in the question, as well as specific values. For value conditions, if the user question specifies a specific value, follow this closely. \\
- Think step by step to find the correct SQL query. \\
\end{tcolorbox}

\subsection{Keyword Extraction}
\label{app:keyword}

The keyword extraction prompt applied is adapted from CHESS \cite{chess}, where various few-shot examples are provided to guide the language model. We adopt their core structure, while removing instructions on value extraction, which CHESS uses to find relevant schema context based on specific values referenced in user questions. We do not explore additional indexing of database values, as this can lead to extreme cost and complexity for massive enterprise datasets with terabytes of data and constantly changing values.

\begin{tcolorbox}[
    breakable,
    colback=gray!10,
    colframe=gray!10,
    left=6pt,
    right=6pt,
    top=6pt,
    bottom=6pt
]
You will be provided with a user question that can be answered by querying some database system. Your objective is to analyze the question to identify and extract keywords and keyphrases which might help indicate what parts of the database schema to use. These elements are crucial for understanding the core components of the question provided. This process involves recognizing and isolating significant terms and phrases that could be instrumental in formulating searches or queries related to the posed question. You should focus on entities such as column or table names that may be referenced, as well as descriptions of what these are. Do not focus on specific column values that may be referenced. \\

\#\#\# Instructions \\

- Read the Question Carefully: Understand the primary focus and specific details of the question. Look for any named entities types (such as organization, location, etc.), technical terms, and other phrases that encapsulate important aspects of the inquiry. \\

- Keywords: Single words that capture essential aspects of the question or hint. \\
- Keyphrases: Short phrases or named entity types that represent specific concepts, locations, organizations, or other significant details. \\

Example 1: \\
Question: "What is the annual revenue of Acme Corp in the United States for 2022? Focus on financial reports and U.S. market performance for the fiscal year 2022." \\

["annual revenue", "corporations", "country", "year", "financial reports", "U.S. market performance", "fiscal year", "corporate revenues"] \\

Example 2: \\
Question: "In the Winter and Summer Olympics of 1988, which game has the most number of competitors? Find the difference of the number of competitors between the two games. the most number of competitors refer to MAX(COUNT(person\_id)); SUBTRACT(COUNT(person\_id where games\_name = '1988 Summer'), COUNT(person\_id where games\_name = '1988 Winter'));" \\

["olympic games", "competitors", "number of competitors", "person\_id", "games", "games\_name", "competitors competing in olympic games"] \\

Example 3: \\
Question: "How many Men's 200 Metres Freestyle events did Ian James Thorpe compete in? Men's 200 Metres Freestyle events refer to event\_name = 'Swimming Men''s 200 metres Freestyle'; events compete in refers to event\_id;" \\

["events", "event\_id", "event\_name", "compete in", "competitors", "competitive games", "competitors competing in events"] \\

Example 4: \\
Question: "List the infant mortality of country with the least Amerindian." \\

["mortality rate", "infant mortality rate", "country", "ethnicity", "population", "infant mortality"] \\

Example 5:  \\
Question: "What are the first names of the students who live in Haiti permanently or have the cell phone number 09700166582?" \\

['students', 'first names', 'country', 'permanent address', 'cell phone number', 'contact information', 'student location', 'student contact details'] \\

\#\#\# Task \\

Given the following question, identify and list all relevant keywords and keyphrases which may indicate which parts of a database schema might be necessary to answer the user question. \\

Question: \{QUESTION\} \\

Please provide your findings as a json list, capturing the essence of the question through the identified terms and phrases. \\
Only output the json list with no explanations.
\end{tcolorbox}

\subsection{Table Description Synthesis Prompt}
\label{app:table_synth}

Table descriptions are synthesized using the below LLM prompt, where table schemas are generated following \ref{app:schema_format}.

\begin{tcolorbox}[
    breakable,
    colback=gray!10,
    colframe=gray!10,
    left=6pt,
    right=6pt,
    top=6pt,
    bottom=6pt
]

You are a database expert creating semantic table descriptions for a text-to-SQL retrieval system. \\

\#\#\# TASK \\
Generate a concise, high-level description of the provided database table that captures its semantic purpose and usage patterns. This description will be embedded in a structured XML format and used for retrieving relevant tables when processing natural language questions. \\

\#\#\# DATABASE SCHEMA \\
\{TABLE\_SCHEMA\} \\

First, think through what makes this table important, what business concepts it represents, and how users might refer to it in natural language questions. Consider its likely role in the database without listing all columns. \\

Then, create a concise table description that covers: \\
1. The table's main purpose and real-world concept it represents \\
2. Its context within the broader database domain \\
3. Typical query patterns or business questions it helps answer \\
4. Key relationships with other tables (if any) \\
5. Alternative terms users might use when referring to this table \\

Keep your description under 150 words, focusing on semantic meaning rather than technical details. For tables with many columns, focus on the overall table purpose and categories of data rather than describing individual columns. \\

Format your response as follows: \\
<thinking> \\
Your analysis of the table and reasoning about its purpose, usage, and significance. \\
</thinking> \\

<description> \\
Paragraph 1: Purpose and domain context of the table in 2-3 sentences. \\
Paragraph 2: Usage patterns and typical questions this table answers in 2-3 sentences. \\
Paragraph 3: Key relationships with other tables in 1-2 sentences (if applicable). \\
Alternative terms: comma-separated list of 3-6 alternative phrases users might use. \\
</description> \\

\end{tcolorbox}

\section{Schema Format}
\label{app:schema_format}

Below is the schema format used for all LLM operations (table prediction and SQL generation). We adopt a format similar to M-Schema \cite{mschema}, with XML tags substituted for better alignment with Anthropic Claude models. For any table-level entities we automatically add $table\_name$ from the metadata, and for all column-level entities we add in $column\_name$ and $data\_type$. We have a separate XML section for $table\_description$, as this is the only table-level contextual entity used, although other sections can be added if additional information types exist. For column-level context entities, we map from a short description of the context type to the value, with the specific example below shown for the BIRD dataset. Foreign keys are added at the database-level when available.

\begin{tcolorbox}[
    breakable,
    colback=gray!10,
    colframe=gray!10,
    left=6pt,
    right=6pt,
    top=6pt,
    bottom=6pt
]
<db>\$db\_id \\

<table>\$table\_name \\
<desc> \\
\$table\_description \\
</desc> \\

<schema> \\
(\$column\_name:\$data\_type, \$column\_alias, Column description: \$column\_description, Value description: \$value\_description), \\
(\$column\_name:\$data\_type, \$column\_alias, Column description: \$column\_description, Value description: \$value\_description), \\
... \\
</schema> \\
</table> \\

<table>\$table\_name \\
<desc> \\
\$table\_description \\
</desc> \\

<schema> \\
(\$column\_name:\$data\_type, ...), \\
(\$column\_name:\$data\_type, ...), \\
... \\
</schema> \\
</table> \\

... \\

<foreign\_keys> \\
\$table\_name.\$column\_name=\$table\_name.\$column\_name \\
... \\
</foreign\_keys> \\

</db> \\

<db>\$db\_id \\

... \\

</db>
\end{tcolorbox}

\section{Analysis of Conflicting Information Overlap}
\label{app:conflict_analysis}

Below we highlight some common causes for low table-level recall at low $N$ values due to highly overlapping table/column information across full datasets. We observe that while performance is affected at low $N$, this effect is eliminated at higher $N$ values, while still reducing overall database schema to a small fraction (e.g. < 3\% of $E_{\Lambda_T}$ and < 1\% of $E_{\Lambda_C}$ at $N=50$) of the original schema.

\subsubsection{Example 1 (Spider)}
\textbf{Question}: Find the districts in which there are both shops selling less than 3000 products and shops selling more than 10000 products. \\
\textbf{Ground truth tables}: \{'employee\_hire\_evaluation.shop'\} \\
\textbf{Ground truth columns}: \{'employee\_hire\_evaluation.shop.district', 'employee\_hire\_evaluation.\\shop.number\_products'\}

\textbf{RASL-Extracted Keywords}: [districts, shops, products, product quantity, shop inventory, retail locations]

\textbf{Finding}: 19 tables contain the word 'products', 6 contain 'shop', and 8 contain 'store'. Similarly, 51 columns contain 'product', 14 contain 'shop', and 11 contain 'district'. While entropy-guided relevance calibration down-weights keywords such as column-level 'product', the effect is less pronounced on high overlap across only 5-15 entities, requiring larger top-$N$ values to load all relevant context.

\subsubsection{Example 2 (BIRD)}
\textbf{Question}:  Where is Amy Firth's hometown? hometown refers to city, county, state \\
\textbf{Ground truth tables}: \{'student\_club.member', 'student\_club.zip\_code'\}\\
\textbf{Ground truth columns}: \{'student\_club.member.first\_name', 'student\_club.member.last\_name', 'student\_club.member.zip', \\ 'student\_club.zip\_code.city', 'student\_club.zip\_code.county', 'student\_club.zip\_code.state', 'student\_club.zip\_code.zip\_code'\} \\

\textbf{RASL-Extracted Keywords}: ['hometown', 'city', 'county', 'state', 'location', 'residence', 'person details'] \\

\textbf{Findings}: 7 tables contain the word 'city', 7 contain 'location', and 5 contain 'state. Similarly, 19 columns contain 'city', 16 contain 'state', and 13 contain 'location'. Due to this highly overlapping information of similar data across different databases, it can be difficult to retrieve top-$N$ context entities at low $N$ values, whereas at higher $N$ values (e.g. 20-50), all sufficient context can be included for schema reasoning.

\section{Baseline Method Implementation Details}
\label{app:baseline_details}

For $Recall@5$ and $Recall@15$ metrics, we directly adopt the metrics reported by DBCopilot \cite{dbcopilot} on external benchmarks, which serializes table schemas into documents containing table names, column names, and column context for use in BM25, SXFMR, and DTR. BM25 uses Okapi BM25 with the two adjustable parameters optimized over training, and SXFMR uses \textit{all-mpnet-base-v2} \cite{mpnet}. CRUSH uses \textit{gpt-3.5-turbo-0125} \cite{openai} for schema hallucination and adopts the same BM25 and SXFMR settings. DTR and DBCopilot are fine-tuned using $1 \times 10^5$ question-SQL pairs, which are synthesized over the target databases using the process detailed in \cite{dbcopilot}.

For ablation study reproductions, we leverage the same settings for BM25 and SXFMR, where we find BM25 works best on training samples using shingles of $K=4$ for Spider and $K=5$ for BIRD and Fiben, with word-level indexing performing poorly across all datasets. For CRUSH, the same settings are used for BM25 and SXFMR, but we leverage Anthropic Calude 3.5 Sonnet-v2 \cite{claude_sonnet} for schema hallucination due to access constraints for OpenAI models.

\section{Cost Analysis}
\label{app:cost_analysis}

A detailed input token cost analysis of RASL versus baseline methods for end-to-end SQL generation is provided in Table~\ref{tab:cost_analysis}. We adopt standard on-demand commercial API pricing provided by Amazon Bedrock \cite{bedrock_pricing}, which charges \$0.0008 per 1,000 input tokens for Anthropic Claude 3.5 Haiku, \$0.003 per 1,000 input tokens for Anthropic Claude 3.5 Sonnet v2, and \$0.0001 per 1,000 input tokens for Cohere Embed 3 English. Cost metrics are broken down by retrieval, LLM prompt, and schema components, with all values calculated per 100 questions to improve readability. LLM prompt includes both the table prediction and SQL generation prompt steps for RASL and only SQL generation for baselines, with the schema prompt component removed to isolate variable costs.

The results demonstrate RASL's superior cost scaling characteristics despite retrieval overhead. While RASL incurs additional costs for retrieval (\$0.03 per 100 queries) and more complex prompting (\$0.39-0.41 vs \$0.10-0.11 for baselines), its key advantage lies in maintaining constant costs as the number of tables increases. Most notably, RASL costs remain identical at N=15 and N=30, while baseline costs scale linearly with the number of included tables. For Spider, RASL costs \$0.80 per 100 queries at both N=15 and N=30, compared to baselines ranging from \$0.45-0.52 at N=15 but increasing to \$0.81-0.93 at N=30. For BIRD, RASL costs \$1.85 per 100 queries versus \$1.32-1.78 for baselines at N=15, but baselines increase substantially to \$2.68-3.22 at N=30. This constant cost scaling, combined with superior accuracy performance, makes RASL particularly attractive for enterprise deployments where database catalogs continue to grow over time.

We believe further optimizations could provide additional cost benefits. Prompt reduction techniques could minimize the fixed prompt overhead (\$0.39-0.41 per 100 queries), while evaluations of lighter-weight models for table prediction (e.g., using Claude Haiku instead of Sonnet) represent promising avenues for cost reduction that warrant investigation. These optimizations represent important directions for future work to enhance RASL's cost-effectiveness in resource-constrained environments.

\begin{table*}[h]
\centering
\small
\begin{tabular}{l|ccc|c|ccc|c}
\hline
\multirow{2}{*}{Method} & \multicolumn{4}{c|}{Spider (per 100 queries)} & \multicolumn{4}{c}{BIRD (per 100 queries)} \\
\cline{2-9}
 & Schema & Retrieval & Prompt & Total & Schema & Retrieval & Prompt & Total \\
 & (N=15/30) & & & (N=15/30) & (N=15/30) & & & (N=15/30) \\
\hline
BM25 & \$0.39/0.80 & \$0.00 & \$0.10 & \$0.49/0.90 & \$1.22/2.57 & \$0.00 & \$0.11 & \$1.32/2.68 \\
SXFMR & \$0.35/0.71 & \$0.00 & \$0.10 & \$0.45/0.81 & \$1.50/2.70 & \$0.00 & \$0.11 & \$1.60/2.80 \\
CRUSH\_BM25 & \$0.42/0.83 & \$0.00 & \$0.10 & \$0.52/0.93 & \$1.54/2.99 & \$0.00 & \$0.11 & \$1.65/3.10 \\
CRUSH\_SXFMR & \$0.41/0.82 & \$0.00 & \$0.10 & \$0.51/0.92 & \$1.68/3.12 & \$0.00 & \$0.11 & \$1.78/3.22 \\
\hline
RASL\_full & \$0.38/0.38 & \$0.03 & \$0.39 & \$0.80/0.80 & \$1.41/1.41 & \$0.03 & \$0.41 & \$1.85/1.85 \\
\hline
\end{tabular}
\caption{Cost breakdown per 100 queries (USD) showing schema, retrieval, and prompt costs.}
\label{tab:cost_analysis}
\end{table*}

\end{document}